\documentclass[sigconf]{aamas} 
\usepackage{preamble}

\setcopyright{ifaamas}
\acmConference[AAMAS '23]{Proc.\@ of the 22nd International Conference
on Autonomous Agents and Multiagent Systems (AAMAS 2023)}{May 29 -- June 2, 2023}
{London, United Kingdom}{A.~Ricci, W.~Yeoh, N.~Agmon, B.~An (eds.)}
\copyrightyear{2023}
\acmYear{2023}
\acmDOI{}
\acmPrice{}
\acmISBN{}

\acmSubmissionID{964}

\title{Automated Task-Time Interventions to Improve Teamwork\\using Imitation Learning}

\author{Sangwon Seo}
\affiliation{
  \institution{Rice University}
  \city{Houston}
  \country{TX, United States}}
\email{Sangwon.Seo@rice.edu}

\author{Bing Han}
\affiliation{
  \institution{Rice University}
  \city{Houston}
  \country{TX, United States}}
\email{bh39@rice.edu}

\author{Vaibhav Unhelkar}
\affiliation{
  \institution{Rice University}
  \city{Houston}
  \country{TX, United States}}
\email{Vaibhav.Unhelkar@rice.edu}

\begin{document}

\pagestyle{fancy}
\fancyhead{}

%%% Use this environment to specify a short abstract for your paper.

\begin{abstract}
Effective human-human and human-autonomy teamwork is critical but often challenging to perfect. The challenge is particularly relevant in time-critical domains, such as healthcare and disaster response, where the time pressures can make coordination increasingly difficult to achieve and the consequences of imperfect coordination can be severe. To improve teamwork in these and other domains, we present \attic: an automated intervention approach for improving coordination between team members. Using BTIL, a multi-agent imitation learning algorithm, our approach first learns a generative model of team behavior from past task execution data. Next, it utilizes the learned generative model and team's task objective (shared reward) to algorithmically generate execution-time interventions. We evaluate our approach in synthetic multi-agent teaming scenarios, where team members make decentralized decisions without full observability of the environment. The experiments demonstrate that the automated interventions can successfully improve team performance and shed light on the design of autonomous agents for improving teamwork.
\end{abstract}

%%% The code below was generated by the tool at http://dl.acm.org/ccs.cfm.
%%% Please replace this example with code appropriate for your own paper.


%%% Use this command to specify a few keywords describing your work.
%%% Keywords should be separated by commas.

\maketitle

\newif\ifarxiv
\arxivtrue

\section{Introduction}
The success of human enterprise depends on effective teamwork.
In domains as diverse as manufacturing, disaster response, and healthcare rarely a single agent conducts a task alone; instead, to benefit from complementary expertise of multiple and diverse agents, processes are almost always multi-agent and collaborative.
In certain situations the teaming context is well defined, while in others the teamwork might be loosely-coupled \cite{grosz1996collaborative,ling2003mip,amir2016mip,oliehoek2016concise}.
Further, real-world teams may be composed of humans alone or, with ongoing advances in artificial intelligence, include robots and autonomous agents.
Irrespective of the teaming context or composition, teamwork is often challenging to perfect.
Imperfect coordination among team members can result in unsatisfactory performance or failures in collaborative tasks \cite{xu2013teaming,leach2009assessing}. 
The effect of these imperfections can be particularly severe in time- and life-critical domains, such as disaster response and healthcare, where the time pressures can make coordination increasingly difficult to achieve.
\ifarxiv
\blfootnote{This article is an extended version of an identically-titled paper accepted at the \textit{International Conference on Autonomous Agents and Multiagent Systems (AAMAS 2023)}.}
\else
\blfootnote{An extended version of this paper, which includes supplementary material mentioned in the text, is available at \url{http://tiny.cc/tic-supplement}}
\fi

Recognizing the ubiquity and adverse consequences of imperfect coordination, there is a long and growing body of research on studying and improving teamwork in multiple disciplines.
For instance researchers in team science, by adopting a human factors perspective, have developed theory and methods for team training and assessments \cite{van2011team,salas2013developing,neily2010association,britton2017assessing,macalpine2017evaluating,dias2021dissecting, salas2018science}.
Concurrently, artificial intelligence community has developed models and algorithms for perfecting multi-agent systems \cite{grosz1996collaborative, tambe2005conflicts, oliehoek2016concise, mirsky2022survey}.
While methods and techniques differ across research communities, not least due to the nuances of different teaming contexts, certain shared insights for improving teamwork exist.
For instance, team members must maintain situational awareness of the task context and mental models of their team members and, by leveraging this situational awareness, adapt their plans to achieve and maintain coordination.

In practice, however, realizing this insight is challenging for team members without appropriate interventions.
In many collaborative contexts, agents need to act under partial observability of the task context as well as that of belief, desires, and intentions of their team members \cite{oliehoek2016concise}.
This partial observability makes it challenging for team members to maintain situation awareness and correctly adapt to maintain shared plans.
For this reason, in some domains such as team sports, a team employs a coach who has a better sight of the task environment and can intervene to maintain team coordination \cite{tuyls2021game,brady2022ai}.
In most other domains, however, resource constraints make it difficult to find an expert who can oversee a team and intervene during each task execution.
Instead, teams adopt team science-inspired methods to improve coordination. 
For instance, successful teams participate in training and debriefing sessions \cite{van2011team,salas2013developing,neily2010association}.
They may also conduct self-diagnostic surveys to assess their teamwork \cite{britton2017assessing,macalpine2017evaluating,dias2021dissecting}.
However, these methods are either \textit{ex ante} or \textit{post hoc} and cannot provide task-time interventions.

To complement these methods, in this work, we consider the \textit{problem of algorithmically generating task-time interventions to improve teamwork}.
As reviewed in \autoref{sec:related-work}, there is growing interest in developing artificial agents to improve human decision-making.
However, to our knowledge, this work is the first to explore feasibility of algorithmically generated \textit{task-time} interventions to improve teamwork.
To solve this problem, we provide the algorithm Task-time Interventions for Improving Collaboration (\attic) that builds upon recent advances in multi-agent imitation learning \cite{seo2022semi} and, given a Markovian model of the collaborative task \cite{oliehoek2016concise} and demonstrations, generates automated task-time interventions.

\begin{figure}[t]
  \centering
  \begin{subfigure}[t]{0.48\linewidth}
      \centering
      \includegraphics[width=\textwidth, frame]{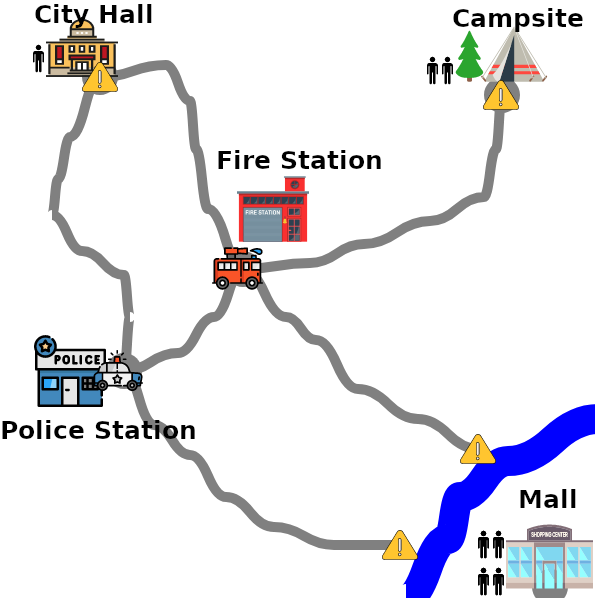}
      \caption{\rescue domain}
      \label{fig: rescue map}
  \end{subfigure}
  \hfill
  \begin{subfigure}[t]{0.48\linewidth}
      \centering
      \includegraphics[width=\textwidth, frame]{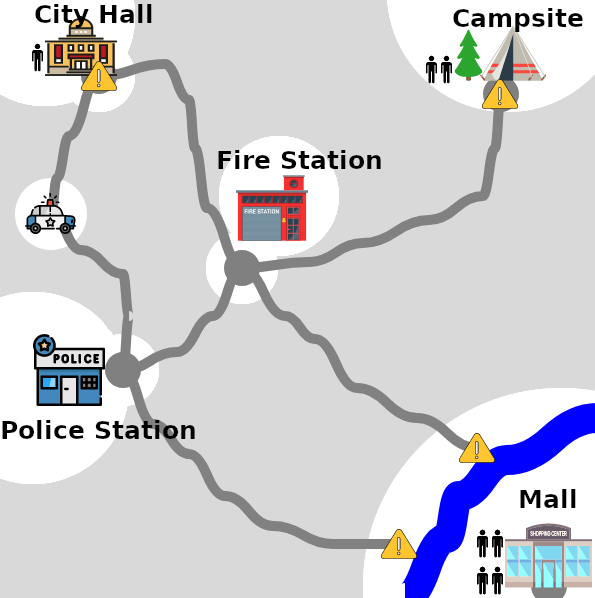}
      \caption{Team member's perspective}
      \label{fig: rescue police sight}
  \end{subfigure}
  \captionsetup{subrefformat=parens}
  \caption{A domain inspired by disaster response, where a team has to rescue people in a decentralized and partially observable environment.}
  \label{fig: rescue}
\end{figure}

\subsection{Running Example}
\label{sec. running example}
To illustrate our problem effectively, we consider a scenario inspired by disaster response \cite{kim2016improving}.
\autoref{fig: rescue map} depicts the scenario, where a team of two members -- a police officer and a firefighter -- is requested to rescue people collaboratively after a flood.
As further shown in \autoref{fig: rescue police sight} (unshaded area), the team members have to operate under partial observability; in particular, each team member can observe their teammate only in their immediate vicinity or at landmarks.
There are four locations (marked with a yellow sign) where people are stranded: a city hall, a campsite, and two broken bridges, each with a different number of people.
While people at city hall and campsite can be rescued just by one agent, repairing a bridge requires both the police officer and firefighter to work together.
As people in the mall can be rescued by repairing just one of the bridges, the team does not have to repair all the bridges.

Each agent can take one of the following actions: \texttt{move} to one of the directions if they are on the road, \texttt{stay} put, or \texttt{rescue} people if they are at a disaster location (yellow sign).
Due to the time-critical nature of the task, in this running example, the team does not have time for prior joint planning and needs to coordinate on-the-fly.
To complete the task effectively (i.e., rescue the most number of people within the time limit), the team members need to coordinate the sites to visit and select cooperative action to realize the shared objective.
As such, the behavior of a coordinated team must depend both on the task context (e.g., location of landmarks, state of the bridge) and team member's mental states (e.g., belief about which sub-activity to pursue next, intent regarding next landmark to visit).

\subsection{Overview of the Proposed Approach}
\label{sec:overview-approach}
As coordination is both critical and challenging in this and other time-critical collaborative tasks, \attic adopts a data-driven approach to generate immediate interventions to improve team coordination during the task execution. 
\attic involves modules to
\begin{enumerate}
    \item[C1.] \textbf{model} each member's behavior and mental states,
    \item[C2.] \textbf{detect} when teamwork is poorly coordinated, and
    \item[C3.] \textbf{intervene} as appropriate to improve the teamwork.
\end{enumerate}

To realize capability C1, \attic requires a behavioral model of each team member that depends on both the task context and team members' mental states (e.g., belief, sub-goals, or intent).
For instance, in our running example, C1 corresponds to prediction of team member's intended sub-goals (e.g., which bridge to repair next) and goal-dependent policies.
Planning or rule-based approaches have been used in the past to model team behavior; however, these techniques either do not account for team member's mental states or, in realistic settings, require significant manual effort to encode behavior for each context \cite{albrecht2018autonomous}.
Recently, imitation learning has garnered popularity in learning predictive models of agent behavior from data \cite{sadigh2016planning,lee2016predicting,osa2018algorithmic}, with growing interest in multi-agent imitation learning \cite{le2017coordinated,song2018multi,wang2021co}.
Informed by successes of these works, we utilize Bayesian Team Imitation Learning (BTIL), a recent multi-agent imitation learning technique to realize C1 \cite{seo2021towards}.
Our choice of the imitation learning algorithm is informed by BTIL's ability to learn behavioral models of teamwork that (a) explicitly depend on team members' mental states, and (b) model team behavior under both perfect and imperfect coordination settings.

During execution of collaborative tasks, poor coordination can result from a variety of factors: misalignment in team member's mental states, imperfect task execution by a particular team member, among others.
For instance, in our running example, the team members may each select a different bridge to repair resulting in imperfect coordination, lost time, and poor task outcomes.
In this work, we focus on imperfect coordination resulting from team member's mental states, which has been identified as a significant factor influencing teamwork \cite{converse1993shared, salas2018science}.
As team member's mental states are not readily observable, solutions are needed to infer them from observed data.
Thus, to realize capability C2 in \attic, we propose a Bayesian approach that estimates at \textit{task time} alignment or lack thereof in team member's mental states.
The proposed Bayesian approach, described in \autoref{sec. inference}, utilizes the team behavioral model learned using C1 and the observed context (state) of the collaborative task as inputs for mental state inference.

Interventions, when appropriate, can improve teamwork.
For instance, in our running example, interventions that help rectify misalignment in team member's mental states (e.g., which bridge to repair next) can improve task outcomes.
At the same time, however, too many interventions can distract teams from their tasks and lead to unintended side effects (e.g., switching costs in human teams) \cite{horvitz2003learning}.
Thus, approaches that can effectively trade-off cost and benefit of interventions are needed.
In \autoref{sec. intervention strategies}, we propose and study multiple intervention strategies to realize C3 in \attic.
The proposed strategies determine when to intervene based on the team's objective, team behavioral model learned via C1, and the estimate derived using C2.
We emphasize development of strategies that generate context-specific interventions and require minimal manual input to estimate the utility of intervention.

We evaluate \attic with four synthetic teaming scenarios, each inspired by real-world applications and existing research \cite{kim2016improving, seo2022semi}.
The experiments demonstrate that the proposed approach \attic can generate automated interventions to improve team performance and can effectively trade-off the cost and benefit of interventions.
These results help establish the feasibility of automated task-time interventions.
Further, the empirical analysis comparing different intervention strategies (C3) of \attic provide guidance for the design of future AI systems to improve human teamwork. 

\section{Related Work}
\label{sec:related-work}
% While few studies have explored automated methods to improve teamwork, 
% Our work relates to the following research fields. 
% \paragraph{Teamwork Assessment}
Our work is informed by a rich body of research on teamwork, both in humanities and artificial intelligence \cite{van2011team,salas2013developing,neily2010association,britton2017assessing,macalpine2017evaluating,dias2021dissecting, salas2018science, grosz1996collaborative, tambe2005conflicts, oliehoek2016concise, mirsky2022survey}.
In this brief review, we discuss approaches for team assessment (relevant to C2) and intervention (relevant to C3).

\paragraph{Methods for Team Assessment}
Researchers in team science have established robust practices for assessing human-only teams \cite{salas2018science,granaasen2019towards,lawson2017computerized}; however, these assessment approaches are post-hoc and do not involve automation.
As AI agents increasingly become part of human teams, recent studies in human-robot interaction suggest several metrics for the teamwork of various types of teams beyond human teams \cite{hoffman2019evaluating,ma2022metrics,norton2022metrics}.
However, to our knowledge, no methods exist to compute these metrics at task time using automated techniques.
Other works present team plan or team intention recognition methods to predict team behavior \cite{yue2015logical,saria2004probabilistic,masato2011agent}, but do not explicitly draw the connection between team intention and team effectiveness. 
\citet{seo2021towards} leverage the concept of shared mental models to recognize ineffective teamwork based on the recognition of each member's mental state.
However, they assume perfect knowledge of team behavior and do not explore the problem of when to intervene to achieve effective teamwork. 

\paragraph{Enhancing Teamwork using AI}
Across domains, there is growing interest in developing artificial agents to improve teamwork \cite{kamar2009incorporating, amir2016mip, kim2016improving, seo2021towards, tuyls2021game, jiang2018learning,foerster2016learning}; however, to our knowledge, our work is the first to explore automated \textit{task-time} interventions in tightly-coupled and time-critical tasks.
For instance, \citet{amir2016mip} present a model of information sharing between the loosely-coupled team, which helps decide when to share information between partners based on the concept of degree of interest.
\citet{kim2016improving} provide an approach to enhance teamwork be developing an intelligent agent to improve human team planning.
Their approach detects the degree of shared understanding by analyzing the dialogues in a meeting before conducting the task. 
\citet{tausczik2013improving} provide real-time language feedback based on the communication patterns of a team. 
% However, their approach is limited to teams with frequent conversations and cannot be applied to teams in non-communicable or communication-challenging domains.
In practice, our approach to generating \textit{task-time} interventions could serve as a useful complement to these existing approaches for generating \textit{planning-time} interventions.
\section{Problem Formulation}
\label{sec:problem}

To address the problem of generating automated task-time interventions for improving teamwork, we first formalize it using existing models of teamwork \cite{oliehoek2016concise}.
Further, we focus on collaborative tasks with a well-defined notion of shared objective.
While our formalism is agnostic to team composition, it is particularly relevant for teams with no hierarchy and at least one human member, i.e., equal partners human teams or human-robot teams.

\subsection{Model of Collaborative Task}
We focus on teams that perform finite-horizon tasks with well-designed objectives in partially observable environments.
To formulate team behavior while taking into account the relevant uncertainties, we model the task as a decentralized partially observable Markov decision process (\decpomdp).
\decpomdp represents a task by a tuple $(n, S, A, \Omega, T_s, O, R, \gamma, h)$, where $n$ is the number of team members, $S$ is the set of task states, $A$ is the set of joint actions $a$ of team members, $\Omega$ is the set of joint observations of team members, $T_s : S \times A \times S \rightarrow [0, 1]$ denotes the state transition probabilities,  $O : \Omega \times A \times S \rightarrow [0, 1]$ denotes the observation probabilities, $R: S \times A \rightarrow \mathbb{R}$ denotes the shared reward, $\gamma$ is the discount factor, and $h$ is the task horizon.
The joint action $a \doteq (a_1, \cdots, a_n)$ can be factorized as $A \doteq \times_i A_i$, where $A_i$ denotes the set of the $i$-th member's actions $a_i \in A_i$.
For more details on \decpomdp, we refer the reader to the text by \citet{oliehoek2016concise}.

\subsection{Model of Team Behavior}
In theory, given the task model, a team can act optimally by computing a decentralized policy using \decpomdp solvers and then executing it in lockstep.
However, such optimal behavior is unrealistic to expect in practice.
First, computing optimal policies for \decpomdp models remains computationally prohibitive for many practical problems, due to which multi-robot teams often employ heuristic policies.
Further, even when an optimal policy can be computed, it is unrealistic to expect team members (particularly humans as well as imperfect robots) to follow such a policy in lockstep and without any execution errors.
In practice, behavior of human team members is often bounded rational and depends on mental states, such as personal preferences or physiological states \cite{gremillion2018estimating, hiatt2017human, simon1990bounded}.
To reflect the dependence of team member's behavior on the mental states, we explicitly model task-relevant mental states as $x_i \in X_i$ and assume them to be unobservable for all team members $j \neq i$.

Borrowing from the Agent Markov Model \cite{unhelkar2019learning}, we represent the behavior of the $i$-th member to depend on the tuple $(X_i, b_{x_i}, T_{x_i}, \pi_i)$, where $X_i$ is the set of their mental states, $b_{x_i}$ is the probability distribution of the mental state at time $0$, $T_{x_i}$ denotes the transition dynamics of the mental state, and $\pi_i$ denotes the member's policy.
Concretely, $T_{x_i}$ denotes the probability of $i$-th team member's mental state at the next step, conditioned on the task history and their current mental state.
Similarly, $\pi_i$ denotes their probability of selecting action $a_i \in A_i$ conditioned on the task history and their current mental state.
While in general the tuple $(X_i, b_{x_i}, T_{x_i}, \pi_i)$ will be unknown for each team member, for tractability, we assume that the set of mental states $X_i$ for each member is known and finite.
Similar to joint actions, we denote $x \doteq (x_1, \cdots, x_n)$, and $X \doteq \times_i X_i$.

\subsection{Problem Statement}
Having identified the models of task and team behavior, we next formalize the problem of interest.
The space of potential interventions to improve teamwork is large.
A coach may intervene a team to convey optimal actions, mental states, plans, among other variables of interest \cite{tuyls2021game}.
In this work, we limit the intervention content to recommended value of mental states $(x)$ which improve team coordination.
We leave the exploration of other types of intervention to future work and, instead, focus on the problem of deciding whether or not to intervene to improve team performance.
We use $z$ to denote the binary decision of intervening in team's task execution.
The variable $c$ denotes the cost per intervention and serves as a simple model for adverse consequences of intervention \cite{horvitz2003learning, macmillan2004communication}.
Formally, the problem assumes as input
\begin{itemize}
    \item the task model $(n, S, A, \Omega, T_s, O, R, \gamma, h)$, 
    \item the set of team member's mental states $(X)$, and 
    \item task execution data $(s^{0:t}, a^{0:t-1})$ up to time $t$.
\end{itemize}
Given these inputs, the problem corresponds to deciding $z_t$ (i.e., whether or not to intervene at time $t$) so as to maximize the objective $J = \sum_{t=0}^h (r_t - c_t z_t) $.
Note that the problem assumes that, similar to a sports coach, the intervention system has full observability of the task state $(s)$ but not that of team members' mental states $(x)$.
The objective reflects the trade-off of maximizing team's cumulative reward, while minimizing the cumulative cost of interventions.

\section{Solution}
\label{sec:solution}
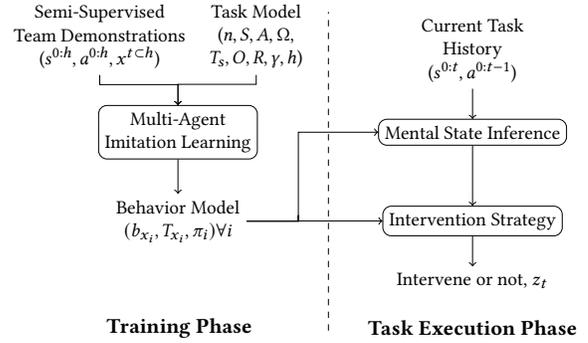
\begin{figure}
    \centering
    \scalebox{0.8}{
    \begin{tikzpicture}
    \node[draw,rounded corners,align=center] (imi) {Multi-Agent\\ \hyperref[sec. imitation learning]{Imitation Learning}};
    \node[draw,rounded corners,right=2cm of imi] (inf) {\hyperref[sec. inference]{Mental State Inference}};
    \node[draw,rounded corners,below=1cm of inf] (intv) {\hyperref[sec. intervention strategies]{Intervention Strategy}};
    \node[below=0.57cm of imi,align=center] (model) {Behavior Model \\ $(b_{x_i}, T_{x_i}, \pi_i) \forall i$};
    
    \node[minimum height=1.2cm,above=0.5cm of imi,xshift=-1.3cm,align=center] (demo) {Semi-Supervised\\Team Demonstrations\\($s^{0:h}, a^{0:h}, x^{t\subset h}$)};
    \node[minimum height=1.2cm,above=0.5cm of imi,xshift=1.3cm,align=center] (mmdp) {Task Model\\($n, S, A, \Omega,$\\$ T_s, O, R, \gamma, h$)};
    \node[above=0.5cm of inf,align=center] (data) {Current Task\\History\\($s^{0:t}, a^{0:t-1}$)};
    \node[below=0.5cm of intv,align=center] (output) {Intervene or not, $z_t$};
    
    \node[below=0.3cm of output,xshift=-0.0cm,align=center] (exec) {\bf \large Task Execution Phase};
    \node[left=2cm of exec,xshift=0.3cm,align=center] (train) {\bf \large Training Phase};

    \path[draw, ->] (demo) |- +(0.2cm,-0.8cm) -| (imi);
    \path[draw, ->] (mmdp) |- +(-0.2cm,-0.8cm) -| (imi);
    \path[draw, ->] (imi) -- (model);
    \path[draw, ->] (model) -| +(2cm, 0.1cm) |- (inf);
    \path[draw, ->] (model)  -| +(2cm, 0.1cm) |- (intv);
    \path[draw, ->] (data) edge (inf);
    \path[draw, ->] (inf) edge (intv);
    \path[draw, ->] (intv) edge (output);
    \draw [dashed] (2.5,-3.3) -- (2.5,2);

    \end{tikzpicture}
    }
    \caption{\fontdimen2\font=0.2em \attic schematic: During the training phase, \attic learns a behavioral model of the team from past execution data. Upon training, when a team performs a task, \attic detects poor coordination and intervenes to improve teamwork.}
    \label{fig. method overview}
\end{figure}

A trivial solution to this problem is to always intervene; however, this solution is likely to be sub-optimal due to the cost of intervention.
Hence, we propose an approach that seeks to generate effective interventions by reasoning about team members' behavior and mental states.
In particular, to derive the solution for the formulated problem, we break down the problem into three sub-problems
\begin{enumerate}
    \item[P1.] \textbf{Model:} Given the task model $(n, S, A, \Omega, T_s, O, R, \gamma, h)$, the set of team member's mental states $(X)$, and partially annotated sequences of team's past $m$ task executions $(s^{0:h}, a^{0:h}, x^{t\subset h})$, \textit{learn a model of each team member's behavior} $(b_{x_i}, T_{x_i}, \pi_i) \forall i$.
    \item[P2.] \textbf{Detect:} Given the inputs of P1, a learned model of each team member's behavior $(b_{x_i}, T_{x_i}, \pi_i) \forall i$, and observable data of on-going task execution $(s^{0:t}, a^{0:t-1})$, \textit{estimate at task-time each team member's mental state} $(x^{t})$.
    \item[P3.] \textbf{Intervene:} Given the inputs of P2 and an estimate of each team member's mental state $(x^{t})$, \textit{decide at task-time whether to intervene} $(z_t)$ so as to maximize the objective $J$.
\end{enumerate}
Our approach includes three key modules, one corresponding to each sub-problem described above.
\autoref{fig. method overview} depicts a schematic of our approach.
As motivated in \autoref{sec:overview-approach}, first, we use multi-agent imitation learning techniques to solve P1.
Next, use the outcome of P1 and a Bayesian inference technique to solve P2.
And, lastly, propose and study a set of intervention strategies to solve P3.

\subsection{Multi-Agent Imitation Learning}
\label{sec. imitation learning}
\newcommand{\JP}{\pi}  % joint policy
\newcommand{\ldls}{\chi_{1:l}}  % labeled dynamic latent states
\newcommand{\udls}{\{x_{m}^{0:h}\}_{m>l}}  % unlabeled dynamic latent states
\newcommand{\JT}{T_x}  % joint transition

To address P1, we observe that the sub-problem of modeling team behavior can be posed as one of imitation learning.
Mathematically, the classical imitation learning setting seeks to learn a single-agent policy $(\pi)$, a function that returns the probability of selecting an action $(a)$ given a task context $(s)$, from a dataset of task executions: $(s, a)$-pairs.
Recently, there is growing interest in addressing the multi-agent version of this problem setting, where the goal is to learn decentralized policy $(\pi_i \forall i)$ of a multi-agent system from demonstrations \cite{le2017coordinated,song2018multi,wang2021co}.
Most multi-agent imitation learning techniques, however, do not explicitly model agent's mental states $(X)$ or learn its temporal dynamics $(T_x)$.

As motivated in \autoref{sec:overview-approach}, multi-agent behavior in teaming context depends not only on the task context $(s)$ but also on a variety of latent decision factors $(x)$, such as preference, sub-goals, or belief.
A few recent papers explicitly include latent factors as independent variables in the behavior model to better account for complex agent behavior \cite{unhelkar2019learning,sharma2018directed,jing2021adversarial,seo2022semi}. 
Among these, only the approach Bayesian Team Imitation Learning (\btil) enables learning of team members' decentralized policies and mental state dynamics from suboptimal demonstrations \cite{seo2022semi}.

Briefly, \btil assumes a Markovian model for the dynamics of team members' mental states $T_{x_i}(x_i'|s, x_i, a, s')$ and policy $\pi_i(a_i|s, x_i)$, and that each agent can fully observe the environment.
Under these assumptions, \btil computes the maximum a posteriori (MAP) estimate of $T_x$ and $\pi$ from past task-execution data.
Since the posterior of $T_x$ and $\pi$ cannot be computed analytically due to latent variables, \btil utilizes mean-field variational inference (MFVI) \cite{hoffman2013stochastic}.
With MFVI, \btil approximates the posterior $p(T_x, \pi | \text{data})$ by maximizing the evidence lower bound(ELBO).
% :
% \begin{align}
%     \small
%     \mathcal{L}(q) := \EX_q \left[ \log \frac{p\left(\JP, \JT, \udls, \text{data}\right)}{q\left(\JP \right)q\left(\JT \right)q\left(\udls\right)} \right] \label{eq.elbo2}
% \end{align}
For more details of \btil, we refer the reader to \cite{seo2022semi}.

\subsection{Learning Models of Team Behavior}
\label{sec:learning-team-model}
The problem setting of \btil corresponds closely to that of P1.
Both settings focus on decentralized team policies, consider mental states and their dynamics, and do not assume perfect coordination in the demonstrations of team behavior.
However, certain key differences also exist.
Most importantly, \btil considers the setting where team members have full observability of the task state, while our problem setting considers partially observable tasks.
As such, the behavior in \btil is modeled using state-dependent policies and not using the general case of history-dependent policies.
% Similarly, the dynamics of mental states $T_x$ are modeled to be Markovian, an assumption which may not hold in the practice.
Learning history-dependent policies of single agents, let alone teams, is a challenging open problem \cite{albrecht2018autonomous}.

However, as we seek to model team behavior from the intervention system's point of view, we can leverage its full observability of the task state during the modeling process.
In particular, we approximate history-dependent policies of team members with state- and belief- (latent state) dependent policy, i.e., $\pi_i \approx \hat{\pi}_i(a_i|s, x_i)$ and $T_{x_i} \approx \hat{T}_{x_i} (x'_i|x_i, a, s')$.
This modeling choice while being an approximation enables us to arrive at a suboptimal but tractable solution to the general problem of modeling team behavior.
Given this modeling choice, the setting of P1 reduces to that of \btil and allowing application of \btil to learn approximations of $(b_{x_i}, T_{x_i}, \pi_i) \forall i$ given inputs to P1.
Application of \btil also enables sample- and label-efficient learning, a desirable attribute in practice.

Our implementation of \btil adopts a modified forward and backward message passing algorithm, which is summarized in the Appendix, by observing that message passing for each agent can be decoupled.
This modification reduces the time complexity of the algorithm from $O(h|X|^{2n})$ to $O(nh|X|^2)$, where $n$ denotes the size of team, $h$ is the task horizon, and $|X|$ denotes the cardinality of set of mental states.
We conclude this section by reiterating that application of \btil in \attic results from an approximate but, as evidenced in our experiments, useful modeling choice.
Learning history-dependent behavior under partial observability remains an exciting open area of future work.
As our approach for generating intervention is modular, its future iterations can adopt advances in partially observable imitation learning to better learn the team model and further improve the quality of interventions.

\subsection{Inferring Team Mental States}
\label{sec. inference}
Leveraging the model learned in P1, we develop a Bayesian inference algorithm to estimate the latent states $(x)$ of the team.
To develop this algorithm, we observe that P2 corresponds to a non-linear multi-agent filtering problem when there are \textit{no} interventions.
Thus, for the case of no interventions, we can formulate the following forward messages to iteratively compute the distribution of mental states at each time step $p(x_i^t | s^{0:t}, a^{0:t-1})$:
\begin{align} \label{eq:infer-no-intervention}
    F(t, x_i &= j)\propto p(x_i^t\myeq{}j, s^{0:t}, a^{0:t-1}) \\
    &= \sum_{k} F(t-1, k) \hat{T}_{x_i}(x_i^t|x_i^{t-1}, a^{t-1}, s^t) \hat{\pi}_i(a_i^{t-1}|s^{t-1}, x_i^{t-1}) \nonumber \\
    F(0, j) &= b_{x_i} (x_i^0\myeq{}j|s^0). \nonumber
    % \pi_i (a_i^0|s^0, x_i^0\myeq{}j)
\end{align}

However, when a team receives intervention, the mental state dynamics $(\hat{T}_{x_i})$ will be affected due to the intervention and thus require modifications to the inference algorithm.
Recall that the problem setting focuses on intervention that convey a set of recommended mental states $(x)$ to the team.
In practice, it is up to each team member's whether to accept the intervention system's recommendation or not; in this work, to simplify our analysis, we assume that each team member accepts the recommendation with probability $p_a$.
Under this assumption, we next update the inference algorithm for the case when an intervention is provided at time $t_{int}$, i.e., $z_{t_{int}} = 1$.
Even in the case of intervention, the algorithm follows an identical structure to \autoref{eq:infer-no-intervention}.
However, the iterative computation of forward messages is updated as follows when $z_{t_{int}} = 1$,

\begin{align}  \label{eq:infer-after-intervention}
    F''(t, x_i &= j) \propto p(x_i^t\myeq{}j, s^{t_{int}+1:t}, a^{t_{int}:t-1} | s^{0:t_{int}}, a^{0:t_{int}-1})\\
        &= \sum_{k} F'(t-1, k) \hat{T}_{x_i}(x_i^t|x_i^{t-1}, a^{t-1}, s^t) \hat{\pi}_i(a_i^{t-1}|s_,^{t-1} x_i^{t-1}), \nonumber\\
    F'(t, x_i) &= \begin{cases}
        F''(t, x_i), \hfill \text{if} \quad t > t_{int}\\
        (1- p_a) p(x_i^t| s^{0:t}, a^{0:t-1}) + p_a \mathds{1} (x_i^t = x_{int, i}), t = t_{int} \nonumber\\
    \end{cases}
    % F'(t_{int}, x_i = j) &= (1- p_a) \cdot p(x_i^t\myeq{}j| s^{0:t_{int}}, a^{0:t_{int}}) + p_a \cdot \mathds{1} (j = x_{int})
\end{align}
where $x_{int, i}$ is the mental state recommendation generated by solving sub-problem P3.
Having computed the distribution of $x$ using  Eqs.~\ref{eq:infer-no-intervention}-\ref{eq:infer-after-intervention}, a point estimate can be obtained by computing the mode of the distribution: $\hat{x}_i = \arg\max_{x_i} p(x_i | s^{0:t}, a^{0:t-1})  \forall i.$

\subsection{Detecting Compatible Mental States}
\label{sec. compatibility}
Having estimated the team member's latent mental states, before intervening, a natural question to ask is whether the team is already coordinated.
To formalize this question, we consider the problem of learning the mental state compatibility function $g(x|s): X  \times S \rightarrow \mathbb{R}$ that takes task context $(s)$ and team members' mental states $(x)$ as input and returns the degree of compatibility.
Observe that the compatibility function is context-dependent, as the same shared preferences may be compatible in some context and incompatible in others.
Further, we consider \textit{degree} of compatibility as opposed to a binary notion, as some mental state combinations might be better than others for the team's performance on the task.

In general, deriving $g(\cdot)$ is challenging. 
Multiple configuration of mental states, which in essence translate to preferences over joint policies or shared plans, can represent perfect coordination.
Further, the answer to the question depends on the task specifications (shared reward) and team members' behavior.
A resource-intensive approach to arrive $g(\cdot)$ is to involve a domain expert who can provide specification of the compatible combinations of mental states.
However, this approach is impractical due to the context-dependent and team-dependent nature of compatibility; the number of states are prohibitively large in practice to allow for manual specification.

Hence, in preparation for our solution for P3, we present a simple metric to approximate $g(\cdot)$.
Given the task model and Markovian model of team behavior learned in \autoref{sec:learning-team-model}, we define $V_\pi(s, x)$ as
\begin{align}
    V_{\pi}(s, x) &= E_{\pi, T_x} \left[\sum_{k=0}^\infty \gamma^k R_{t+k+1}|s^t=s, x^t=x \right]  \label{eq. cumulative reward}
\end{align}
where $R$ is the task reward. 
Observe that $V$ corresponds to the expected cumulative discounted reward (or the V value) of the centralized and fully observable version of the \decpomdp task model conditioned on the learned team policy $\hat{\pi}$.
Several methods exist for computing $V$ values for the fully observable setting, which scale to large problems and are straightforward to implement \cite{puterman2014markov, sutton2018reinforcement}.
The $V$ value depends on the task model, team behavior, current task context and team mental states.
Further, it is straightforward to compute and reflects the impact of team mental states $(x)$ on the team's objective (expected discounted cumulative reward).
As $V_\pi$ fulfills all the feature required of a useful compatibility metric, we use it to estimate $g(x | s) \approx V_\pi (s, x)$.
Similarly, we can also derive the most compatible combination of team members' mental states (i.e., shared preferences which help achieve the perfect coordination) as $x_{int} = \arg\max_{x} V_\pi (s, x)$.

\subsection{Generating Effective Interventions}
\label{sec. intervention strategies}
With an approach to compute compatibility of team members' mental states, we discuss the final component of \attic: strategies to resolve the cost and benefit trade-off of intervention.
While we seek optimal intervention strategies, computationally solving P3 and deriving them is intractable. 
Hence, we propose a suite of heuristic strategies to compute $z_t$ given the problem inputs.
The proposed heuristics can be categorized as deterministic or stochastic, based on their utilization of inference derived in \autoref{sec. inference}.

\subsubsection{Deterministic Strategies}
The first set of strategies rely on a point estimate of the mental state $\hat{x}$ and are denoted using intervention functions $f(\hat{x}| s):X\times S \rightarrow \{0, 1\}$, which assume the current task context $(s)$ and estimated mental state $(\hat{x})$ as inputs.
We present two deterministic strategies: rule-based and value-based.

\paragraph{Rule-based}
When domain knowledge is available, we can rely on manual specifications of compatible mental states.
For this strategy, we assume a domain expert provides a set of compatible mental states denoted by $C_s$.
Given this input, the strategy simply corresponds to verifying if the current estimate of mental state is compatible and intervening if it is not:
\begin{align}
    f_{rule} (\hat{x}| s) &= \mathds{1}\left( \hat{x} \notin C_s \right)
\end{align}
If the decision is to intervene, the strategy recommends one of the compatible mental states from $C_s$ to the team.%\footnote{In cases where $|C_s| > 1$, an interesting question is to decide which of the several compatible states to recommend. We leave exploration of this question to future work.}

\paragraph{Value-based}
In cases where domain knowledge is absent, we rely on the estimate of $g(x|s) \approx V_\pi(x,s)$ to compute $z$.
First, we quantify the benefit of providing an intervention:
\begin{align}
    \texttt{regret}(x|s) &= \max_{x'} g(x'|s) - g(x|s) \\
    \texttt{benefit}(x | s) &= \texttt{regret}(x|s) - c.
\end{align}
Given an estimate of $x$, the strategy decides to intervene when \texttt{benefit} of the intervention exceeds a threshold $\delta$:
\begin{align}
    f_{\delta}(\hat{x}| s) = \mathds{1}\left(\texttt{benefit}(\hat{x}|s) > \delta\right)
\end{align}%
$\delta$ is a domain-dependent hyperparameter, which can help tune the average number of interventions generated by the strategy.
If the decision to intervention is positive, the team is recommended the most compatible mental state: $x_{int} = \arg\max_x g (x|s)$.

\subsubsection{Stochastic Strategies}
\newcommand{\thr}{Confidence-based\xspace}
\newcommand{\avg}{Expectation-based\xspace} 
Instead of relying on the point estimate of $x$, the second set of strategies utilize its inferred distribution $p(x)$.
We denote these strategies using intervention function $\Phi(p | s): \Delta \times S \rightarrow \{0, 1\}$, where $\Delta$ denotes the probability simplex.
We present two heuristics to convert any deterministic strategy $f$ (e.g., rule- or value-based) to its probabilistic counterpart $\Phi_f$. 

\paragraph{\thr} 
Denote $\hat{x} = \arg\max_{x} p(x)$. To avoid spurious interventions, the first heuristic, avoids intervening when the confidence in the estimate is lower than a threshold $\theta$:
\begin{align}
    \Phi_{f, \theta} (p | s) =  f(\hat{x}| s) \cdot \mathds{1} \left( p(\hat{x}) > \theta \right) 
\end{align}
% where $\hat{x} = \arg\max_{x} p(x)$.
% If $\theta \leq 0$, $\Phi_{f,\theta}$ reduces to the deterministic strategy $f$.

\paragraph{\avg} 
The second strategy computes a soft decision, by computing the expected value of the deterministic strategy:
\begin{align}
    \Phi_{f, E}(p | s) = \mathds{1}\left( E_p[f(x |s )] > 0.5 \right)
\end{align}
An intervention is made if the soft decision is greater than half.

\section{Experiments}
\label{sec. experiments}
We evaluate our algorithm in four synthetic scenarios -- \movers, \cleanup, \rescue, and \rescuetwo -- each inspired by real-world applications and with large state space \cite{seo2022semi, kim2016improving}.
All domains involve partial observability, require collaboration, and include multiple sub-tasks.
Team members' mental state $(x)$ correspond to their preference over which sub-task to complete next.
To complete the task efficiently, the team needs to coordinates its next sub-tasks.

\begin{figure}[t]
  \centering
  \begin{subfigure}[b]{0.48\linewidth}
      \centering
      \includegraphics[width=0.9\textwidth, frame]{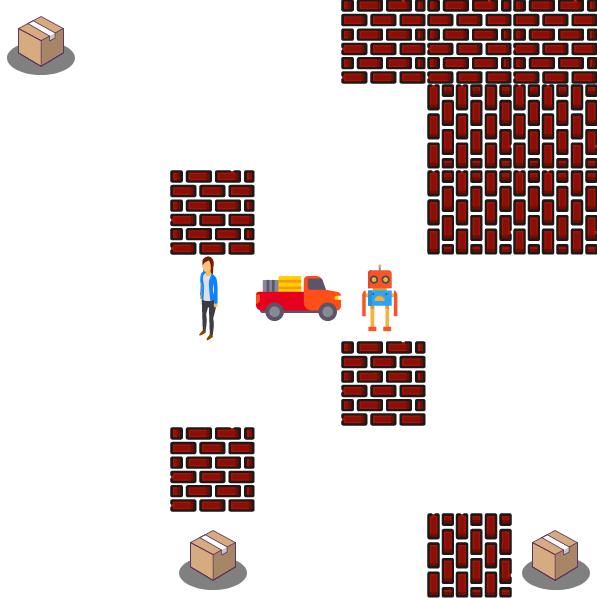}
      \caption{Configuration of \movers}
      \label{fig: movers}
  \end{subfigure}
  \hfill  
  \begin{subfigure}[b]{0.48\linewidth}
      \centering
      \includegraphics[width=0.9\textwidth, frame]{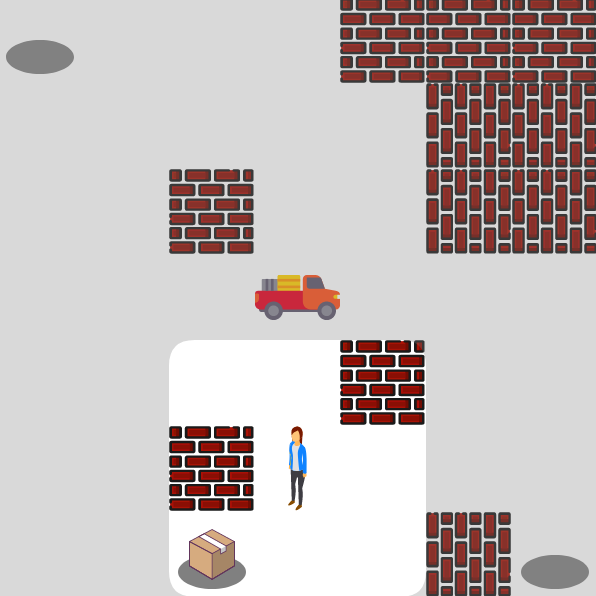}
      \caption{\movers: Alice's perspective}
      \label{fig: movers human sight}
  \end{subfigure}
  \hfill  
  \begin{subfigure}[b]{0.48\linewidth}
      \centering
      \includegraphics[width=0.9\textwidth, frame]{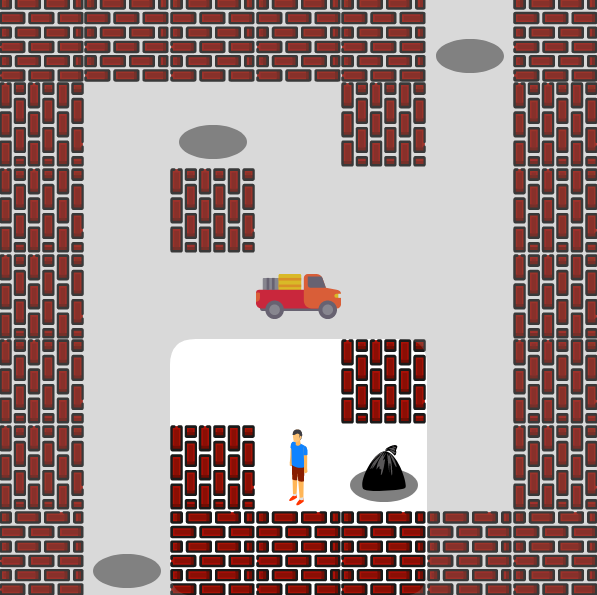}
      \caption{\cleanup: John's perspective}
      \label{fig: cleanup human sight}
  \end{subfigure}
  \hfill  
  \begin{subfigure}[b]{0.48\linewidth}
      \centering
      \includegraphics[width=0.9\textwidth, frame]{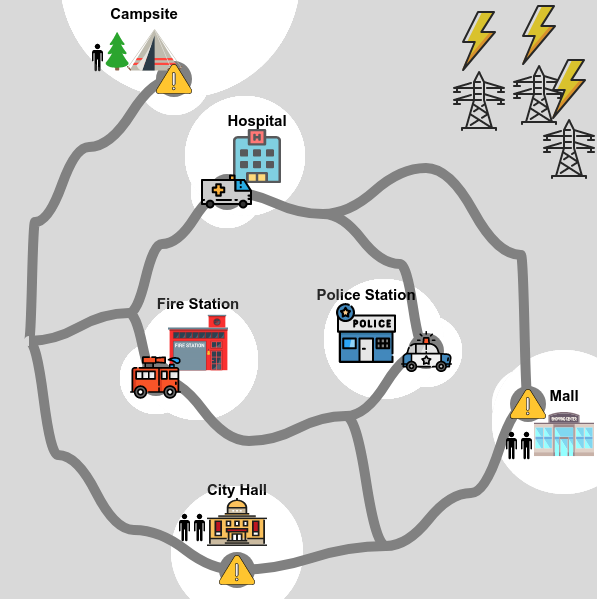}
      \caption{\fontdimen2\font=0.2em \rescuetwo: Police's perspective}
      \label{fig: rescue2 human sight}
  \end{subfigure}

  \captionsetup{subrefformat=parens}
  \caption{\movers, \cleanup and \rescuetwo domains. Agents can observe only part of the environment (unshaded region).}
  \label{fig: domains}
\end{figure}

\subsection{Domains}
\paragraph{\movers}
This domain is similar to \texttt{Movers} domain in \cite{seo2022semi} but adds the challenge of partial observability.
As shown in \autoref{fig: movers}, both agents, Alice and Rob, start their task with a truck in between and have to move the boxes scattered across the space to the truck.
As depicted in \autoref{fig: movers human sight}, however, they can only observe their neighboring (unshaded) region.
The boxes are heavy, requiring two agents to lift them together.
The truck is only accessible while carrying a box, so it is difficult for agents to know where their teammates are at the beginning.
An agent can move in one of four cardinal directions, stay put, or attempt to pick up or drop off a box.
Each team member can select its next sub-task $(x)$ as one of the three pick-up locations or the drop-off location.

\paragraph{\cleanup}
This domain is similar to \texttt{Cleanup} domain in \cite{seo2022semi} but has a different configuration and enforces partial observability to each agent.
A team of two agents, John and Rob, is tasked with moving all the trash bags to the truck.
Unlike \movers, trash bags can only be picked up by one agent. 
The set of an agent's actions is the same as \movers.
While performing the task, an agent may choose one of five mental states: the pickup and drop-off locations.
% : and a total of $|S| \approx 135k$ task states.

\paragraph{\rescue}
This domain implements the running example in \autoref{sec. running example}.
It is an alternate version of \texttt{Emergency Response} domain of \citet{kim2016improving} with the added condition of partial observability.
The number of people rescued via ``rescue'' action is different for each disaster site: one at the city hall, two at the campsite, and four in the mall.
The goal is to rescue as many people as possible within thirty steps.
Each team member can select its next sub-task $(x)$ as one of the four disaster sites.

\paragraph{\rescuetwo}
This domain is similar to \rescue domain but has a different map.
In this domain, a team of three members -- a police officer, a firefighter, and an EMT(Emergency Medical Technician) -- is tasked with rescuing people in the \rescuetwo emergency.
Similar to \rescue, each member can only observe their teammates who are at the same location or at landmarks.
While the person at the campsite can be rescued by one member, to rescue people in the mall and the city hall, at least two members should work together.
Also, the number of people rescued at each site is different: one at the campsite, two in the city hall, and two in the mall.
The team is tasked to recuse as many people as possible within $15$ steps.

\subsection{Data Generation}
To evaluate our approach, we created artificial agents and generated a synthetic dataset by simulating teamwork among these agents.
For each domain, we implemented a transition of the world $T_s$ and designed each agent's ground truth policy $\pi_i$ and their mental state transition $T_{x_i}$. 
Each member's ground truth policy is computed via value iteration by assigning appropriate rewards according to each mental state.
The mental state transition is hand-crafted.
Due to partial observability, agents make decisions based on their own point estimates of the state $\hat{s}$, which may differ from actual task states.
In situations where teammates are not observable, their actions also have no effect on updating an agent's mental state.
Task execution data are generated by randomly sampling the initial mental state $x_i$ at each member, and iteratively sampling each member's action $a_i \sim \pi_i(\cdot |\hat{s}, x_i)$, the next state $s' \sim T_s (\cdot|s, a)$, and the next mental state of each member $x_i' \sim T_{x_i}(\cdot|x_i, \hat{a}, \hat{s}')$ until the completion of the task.
For each domain, we generate $500$ demonstrations for training \btil and $100$ for evaluating our inference algorithm.

\subsection{Evaluation and Metrics}
\paragraph{Inference Algorithm} 
In practice, collecting data of team's task execution $(s^{0:h}, a^{0:h})$ and annotating team member's mental states $x^{0:h}$ is resource intensive.
To empirically assess the sample- and label-complexity of our inference algorithm, we evaluate its performance using different numbers of training data with different supervision of mental states: $150$ demonstrations with $100$\% supervision, $500$ demonstrations with $30$\% supervision (i.e., $150$ labeled demonstrations), and $500$ demonstrations with $100$\% supervision.
For each episode, we compute the accuracy of inference as follows: $\frac{1}{nh}\sum_{i=1}^n \sum_{t=0}^h \mathds{1}(x_i^t = \hat{x}_i^t)$.
Inference accuracy evaluates the effectiveness of our forward prediction algorithm, which utilizes the learned behavioral models, to solve P2. 

\paragraph{Overall Approach} 
Due to its modular nature, \attic can utilize any of the intervention strategies proposed in \autoref{sec. intervention strategies}.
We benchmark multiple versions of \attic, each with different intervention strategy, against two baselines: \textit{no intervention} (i.e., decentralized execution without any intervention) and \textit{centralized policy} (i.e., an optimal team policy which assumes team receives interventions at each step without any consideration of intervention costs).
In these experiments, we conduct $100$ episodic trials for each combination of domain and intervention strategy and report three metrics: cumulative task rewards, the number of interventions, and problem objective $J$.
\attic computes the decision to intervene $z_t$ using observable data of the team's task execution $(s^{0:t}, a^{0:t-1})$ and learned model of team behavior.
To show that \attic can work with less supervision, we use models of team behavior learned using semi-supervised data (i.e., $500$ data with $30$\% supervision).
If \attic decides to intervene, then the team members update their mental state $x$ with the one recommended by the approach (i.e., $p_a = 1$).

For \movers and \cleanup domains, where the reward function reflects time to complete the task, we set cost of intervening as $c = 1$.
For the emergency response domains, where the reward function corresponds to number of people successfully rescued, defining intervention cost is an ill-defined question; for these domains, we set $c=0$ and study the trade-off between task reward and number of interventions.
We also conduct ablation studies to compare the effect of intervention strategy and hyperparameters.
Rule-based strategy is applicable only to those domains where compatibility is easy to define manually.
Hence, we apply the rule-based strategy only to \movers but evaluate the value-based strategy on all domains.
% The set of compatible mental models in \movers is in \autoref{sec. movers compatible}.
We evaluate the proposed deterministic strategies and their two stochastic versions, by varying the hyperparameters: benefit threshold $\delta$ and the certainty threshold $\theta$.

% We present the result along with no intervention setting and perfect coordination setting (i.e. each member executes the optimal, centralized policy).
% Task scores will tell us how effective each automated intervention strategy is and the number of interventions will tell us how efficiently each strategy intervenes.

\subsection{Results}

\begin{figure}[t]
  \centering
  \includegraphics[width=0.8\linewidth]{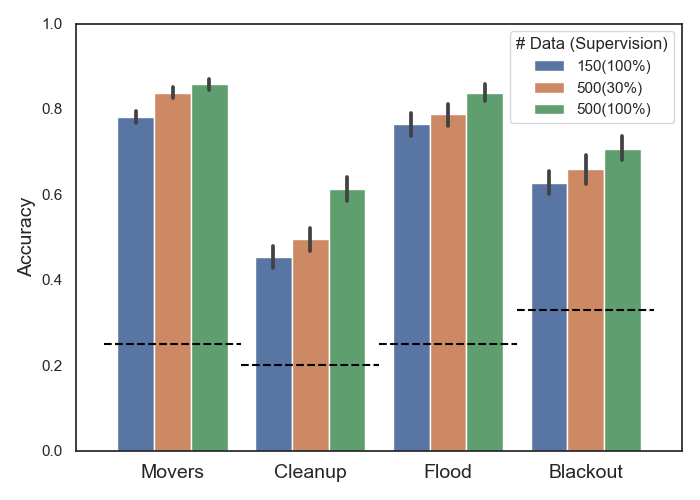}
  \caption{Accuracy of inferring mental states (the dashed black line represents the accuracy of a random guess).}
  \label{fig: inference accuracy}
\end{figure}

\begin{figure}[t]
  \centering
  \includegraphics[width=\linewidth]{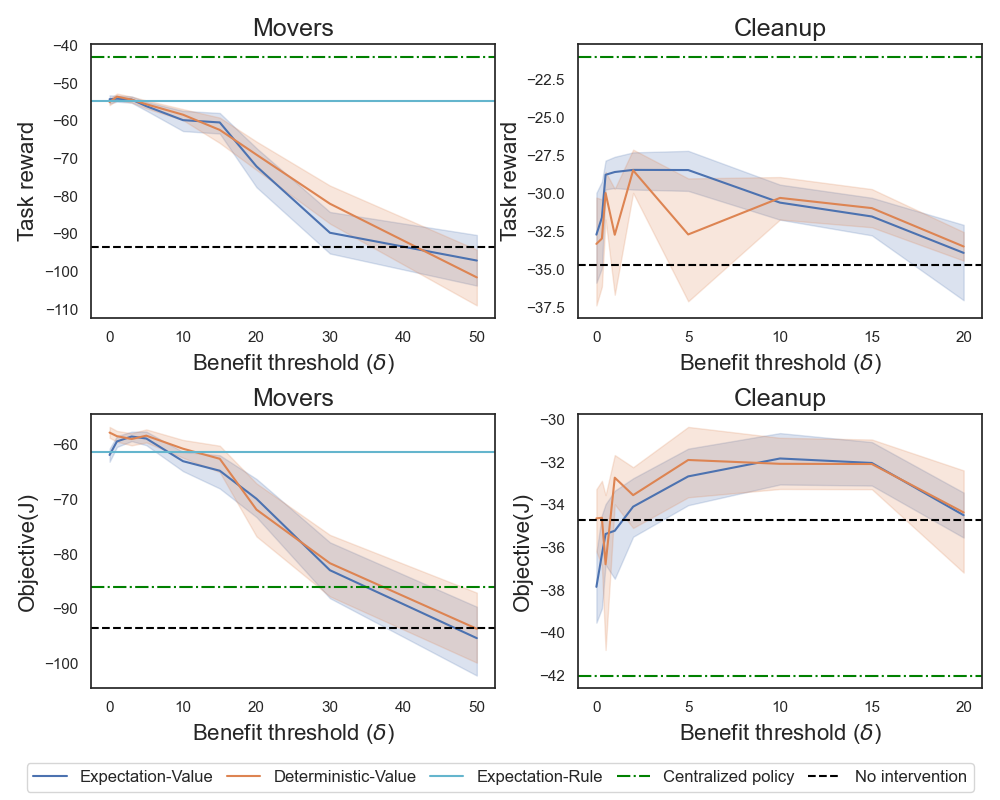}
  \caption{Team's cumulative task reward (top) and the problem objective $J$ (bottom) in \movers and \cleanup domains.}
  \label{fig: boxpush intervention}
\end{figure}

\begin{figure}[t]
  \centering
  \includegraphics[width=\linewidth]{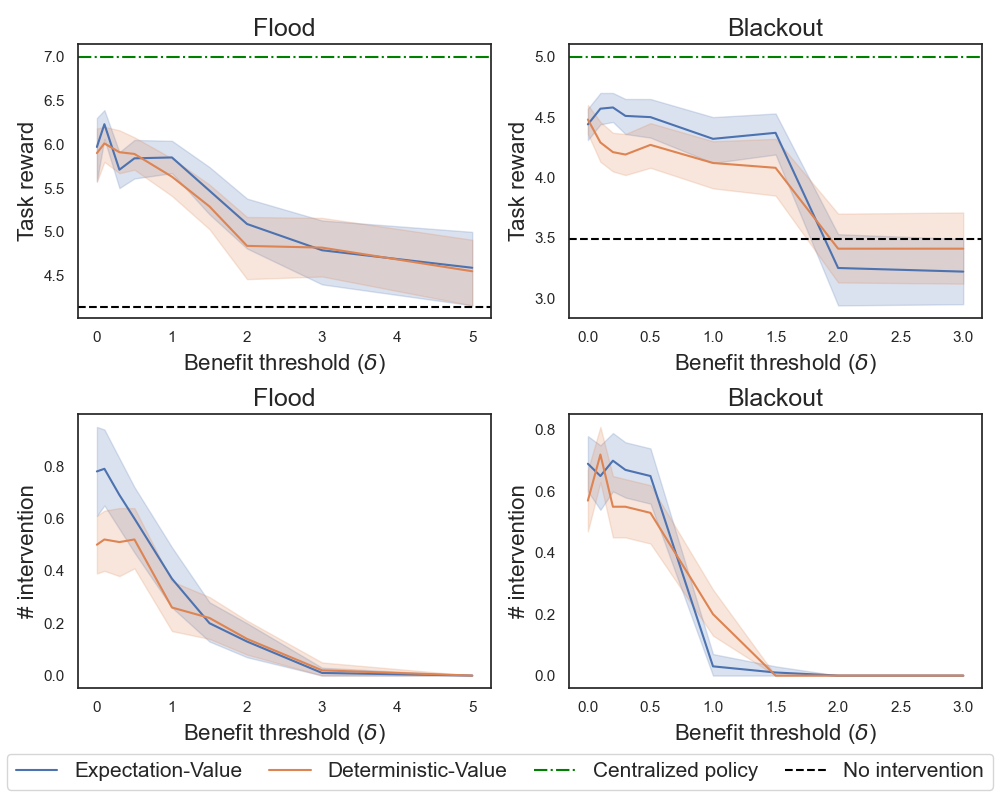}
  \caption{Task reward (top) and the number of interventions (bottom) in \rescue and \rescuetwo domains.}
  \label{fig: rescue intervention}
\end{figure}

\paragraph{Accuracy of Task-time Inference of Mental States}
As shown in \autoref{fig: inference accuracy}, our inference algorithm predicts team members' mental states with over 80\% accuracy in \movers and \rescue, and around 60\% accuracy in \cleanup and \rescuetwo.
In all domains, the algorithm significantly outperforms the probability of random guessing and the performance increases with more data and supervision.
We posit the lower accuracy in \cleanup and \rescuetwo results due to relative lack of training data.
\cleanup has a much larger state space ($\sim$135k) than other domains (\movers: $\sim$40k, \rescue: $\sim$16k, and \rescuetwo: $\sim$74k), resulting in less training data relative to the size of the problem.
\rescuetwo has a much smaller average episode length ($\approx 95$ for \movers, $\approx 15$ for \rescuetwo), resulting in less training data per demonstration.
These results suggest that the proposed approach is capable of learning team behavioral models and inferring mental states with a reasonable accuracy and that, with sufficient training data, the effects of modeling assumptions are not severe.

\paragraph{Utility of Automated Task-time Interventions}
The baselines \textit{no intervention} and \textit{centralized policy} can be viewed as two ends of the spectrum of the solution space.
Unsurprisingly, the \textit{centralized policy} results in the highest task reward.
However, as depicted in \autoref{fig: boxpush intervention} (bottom) and reflected in the problem objective $J$, this high performance comes at a much higher total cost of interventions.
In practice, it is typically infeasible to always intervene either due to environment constraints or intervention overheads.
On the other end of the spectrum, teams that receive no intervention do not incur its costs but also perform poorly.
In comparison and as depicted in Figs.~\ref{fig: boxpush intervention}--\ref{fig: rescue intervention}, \attic (with a suitable choice of hyperparameter $\delta$) successfully helps improve team's task performance for all domains with relatively small number of timely but impactful interventions. 
Encouragingly, \attic is able to generate these interventions despite using noisy inference of team members' mental states.

\begin{figure}[t]
  \centering
  \includegraphics[width=0.8\linewidth]{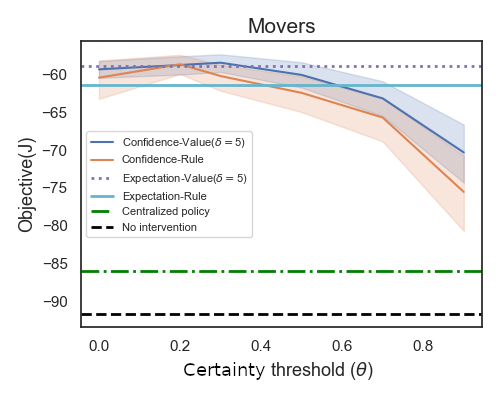}
  \caption{Objective $J$ (with $c=1$) in the \movers domain.}
  \label{fig: score vs theta}
\end{figure}

\paragraph{Effect of Estimating Compatibility}
We observe in \autoref{fig: score vs theta} that the proposed value-based approach to estimating team compatibility $g(\cdot)$ can achieve the same or better performance as the rule-based approach.
This suggests that the value $V_\pi$ defined in \autoref{eq. cumulative reward} is a good proxy for compatibility $g(x|s)$ and enables an objective task-time assessment of teamwork in domains where hand-crafting the compatibility metric may be resource intensive.

\paragraph{Effect of Intervention Strategy and Hyperparameters}
Our ablation experiments comparing different versions of \attic show that even though \attic can improve teamwork, determining the most optimal intervention strategy is difficult due to its dependence on a large number of factors such as task reward, team behavior, inference uncertainties, and intervention costs.
Through Figs.~\ref{fig: boxpush intervention}-\ref{fig: score vs theta}, we also assess the effect of intervention strategy and its hyperparameters $(\delta, \theta)$ on the overall problem objective $(J)$.
We observe that there is no clear winner among proposed intervention strategies and, even with fixed interruption cost, the relationship of objective $J$ with both hyperparameters and number of interventions is non-linear.
Fortunately, our computational approach also offers a mechanism for selecting the appropriate intervention strategy through simulation.
In practical applications of this and similar approaches, we suggest first collecting data of past execution data of real world team's behavior, using this data to learn team model and simulate team behavior (similar to our experiment protocol), and then through simulation conduct a hyperparameter analysis to empirically select the most suitable intervention strategy and its hyperparameters.

\section{Conclusion}
We present \attic, an algorithm for generating automated task-time interventions to improve teamwork.
Using past and current task execution data, \attic detects poor coordination during teamwork and uses heuristics to trade-off costs and benefits of interventions.
We conduct simulation experiments where multi-agent teams perform sequential tasks in partially observable environments and receive interventions generated using \attic.
The results show that automated interventions can successfully improve team performance and provide proof-of-concept for development of automated intervention systems to improve human-human and human-AI teamwork.

% While we discover numerous factors to consider intervening in a team, our work also leaves several unexplored directions and suggests several areas of future work.
% Since we only conducted experiments with simulated teams, the immediate future direction is to explore and model such factors with teams of humans or human-agent.
% Also, we assume the set of mental states is known to derive interpretable and tractable analysis.
% It would be worth exploring non-parametric or unsupervised methods to learn mental states and relate their compatibility to team performance.
% Given that communication is an important part of most teamwork, incorporating communication features into teamwork intervention algorithms is also an important future direction.

%===============================================================================
% The maximum paper length is 8 pages excluding references and acknowledgements, and 10 pages including references and acknowledgements.
\clearpage

\appendix
% \section{Connection to Variable/Shared Autonomy Literature}
% There exists a large body of works to provide the real-time interventions for improving teamwork of human-robot teams \cite{chiou2021mixed,hong2019investigating}. These works address the problem of variable or shared autonomy for robotic systems that may operate autonomously or be directly controlled by human operators. However, in contrast to our approach, they only focus on the specific setting of humans controlling multiple robots. Our approach does not impose such hierarchical structure between team members and treat human team members and robot team members as equal partners. In addition, our approach can be applied to improve performance of human-only teams.

% \section{Model each member’s behavior and mental states (P1)}

% \subsection{Bayesian Team Imitation Learning (BTIL)}
% We utilize a variant of BTIL to solve P1. Employing Bayesian approach, BTIL tries to find $\pi$ and $T_x$ that maximizes the posterior $p(\pi, T_x | data)$. However, computing the posterior is challenging due to unknown variables $\udls$. Thus, using MFVI, BTIL seeks to find an approximate of the posterior by maximizing the following ELBO:
% \begin{align}
%     \mathcal{L}(q) := \EX_q \left[ \log \frac{p\left(\JP, \JT, \udls, \text{data}\right)}{q\left(\JP \right)q\left(\JT \right)q\left(\udls\right)} \right] \label{eq.elbo2}
% \end{align}
% Please refer to \cite{seo2022semi} to see the choice of the priors of $\pi$ and $T_x$ and the actual computation of $q$'s dervied from \autoref{eq.elbo2}.

\section{Modified Forward-Backward Message Passing for Bayesian Team Imitation Learning (BTIL)}
\label{sec. modified btil}

We utilize a variant of BTIL to solve P1 that uses a more efficient forward-backward message passing subroutine.
Observing that the message passing subroutine of BTIL for each agent can be decoupled, we reduce it from $O(h|X|^{2n})$ to $O(nh|X|^2)$ by modifying the messages.
\small
\begin{align*}
    F_i&(t, j) = c^t \cdot p(x_i^t \myeq j, s^{0:t}, a^{0:t})\\
        &=\sum_{k} F_i(t-1, k) T_{x_i}(j | k, s^{t-1}, a^{t-1}, s^t) \pi_i (a^t|j, s^t)\\
    B_i&(t, j) = d^t \cdot p(s^{t+1:h}, a^{t+1:h}|x_i^t=j, s^{0:t}, a^{0:t})\\
    &=\sum_k B_i(t+1, k) T_{x_i}(k|j, s^t, a^t, s^{t+1})\pi_i(a_i^{t+1}|k, s^{t+1})\\
    q&(x_i^t) \propto p(x_i^t, s^{0:h}, a^{0:h}) \propto F_i(t, x_i^t) \cdot B_i(t, x_i^t)\\
    q&(x_i^t, x_i^{t+1}) \propto p(x_i^t, x_i^{t+1}, s^{0:h}, a^{0:h}) \\
    % &\propto p(x_i^t, s^{0:t}, a^{0:t}) p(s^{t+2:h}, a^{t+2:h}|x_i^{t+1}, s^{0:t+1}, a^{0:t+1})\\
    % &\qquad\cdot T_s(s^{t+1}|s^t, a^t)T_{x_i}(x_i^{t+1}|x_i^t, s^t, a^t, s^{t+1})\pi_i(a_i^{t+1}|x_i^{t+1}, s^{t+1})\\
    % &\qquad \cdot p(a_{-i}^{t+1}|s^{0:t+1}, a^{0:t})\\
    &\propto F_i(t, x_i^t) B_i(t+1, x_i^{t+1}) T_{x_i}(x_i^{t+1}|x_i^t, s^t, a^t, s^{t+1})\pi_i(a_i^{t+1}|x_i^{t+1}, s^{t+1})
\end{align*}
\normalsize
where $c$ and $d$ are constants.

Derivation:
\small
\begin{align*}
    \frac{1}{c^t} F_i(t, j) &= p(x_i^t\myeq j, s^{0:t}, a^{0:t}) \\
                &= \sum_k p(x_i^{t-1}\myeq k, s^{0:t-1}, a^{0:t-1})p(j, s^t, a^t|k, s^{0:t-1}, a^{0:t-1})\\
                &= \sum_k \frac{1}{c^{t-1}} F(t-1, k) T_s(s^t|s^{t-1}, a^{t-1}) T_x (j|k, s^{t-1}, a^{t-1}, s^t)\\ 
                &\qquad\qquad\qquad\qquad\cdot\pi_i(a_i^t|j, s^t)p(a_{-i}^t|x_i^{t-1:t}, s^{0:t}, a^{0:t-1}, a_i^t)\\
                &=T_s(s^t|s^{t-1}, a^{t-1})p(a_{-i}^t|s^{0:t},a^{0:t-1})\frac{1}{c^{t-1}}\\
                &\qquad\qquad\cdot\sum_k F_i(t-1, k) T_{x_i}(j|k, s^{t-1}, a^{t-1}, s^t)\pi_i(a_i^t|j, s^t)\\
    c^t &= \frac{c^{t-1}}{T_s(s^t|s^{t-1}, a^{t-1})p(a_{-i}^t|s^{0:t},a^{0:t-1})}, \quad c^0 = 1
\end{align*}

\begin{align*}
    \frac{1}{d^t} B_i(t, j) &= p(s^{t+1:h}, a^{t+1:h}|x_i^t\myeq j, s^{0:t}, a^{0:t})\\
    % &=\sum_k p(s^{t+1:h}, a^{t+1:h}, x_i^{t+1}\myeq k|x_i^t\myeq j, s^{0:t}, a^{0:t})\\
    &=\sum_k p(s^{t+2:h}, a^{t+2:h} |x_i^{t+1}\myeq k, s^{0:t+1}, a^{0:t+1})\\
    &\qquad\qquad\qquad\qquad\cdot p(s^{t+1}, a^{t+1}, k|j, s^{0:t}, a^{0:t})\\
    &=\sum_k \frac{1}{d^{t+1}}B_i(t+1, k) T_s(s^{t+1}|s^t, a^t)T_{x_i}(k|j, s^t, a^t, s^{t+1})\\
    &\qquad\qquad\qquad\cdot\pi_i(a_i^{t+1}|k, s^{t+1})p(a_{-i}^{t+1}|x_i^{t:t+1}, s^{0:t+1}, a^{0:t}, a_i^{t+1})\\
    &=T_x(s^{t+1}|s^t, a^t)p(a_{-i}^{t+1}|a^{0:t}, s^{0:t+1})\frac{1}{d^{t+1}}\\
    &\qquad\qquad\cdot\sum_k B_i(t+1, k) T_{x_i}(k|j, s^t, a^t, s^{t+1})\pi_i(a_i^{t+1}|k, s^{0:t+1})\\
    &d^{t+1} = d^t T_x(s^{t+1}|s^t, a^t)p(a_{-i}^{t+1}|a^{0:t}, s^{0:t+1}), \quad d^h = 1
\end{align*}
\normalsize
Note that $p(a_{-i}^t|x_i^{t-1:t}, s^{0:t}, a^{0:t-1}, a_i^t)=p(a_{-i}^t|s^{0:t},a^{0:t-1})$ due to conditional independence.

\section{Mental State Transition Model ($T_x$) for each domain}
We used ``common sense'' rules to handcraft the mental state transitions. 
\subsection{\movers}
At task start, each agent randomly chooses a target box. Then, they head to the targeting box with probability 1 until reaching a location near enough to see the box. If they are around the destination but do not observe their teammate there, they change their target with 0.1 probability. When they pick up a box, they change their destination to the truck. Once they drop a box, they randomly choose one of the destinations where they think a box exists. 

\subsection{\cleanup}
At task start, each agent randomly chooses a target box. Then, they move to the targeting box unless they realizes the box is already moved or currently held by their teammate. If the box is there, they will pick it up and change the destination to the truck with probability 1. However, if they realize the box is not there (either by checking the drop place or seeing their teammate is currently carrying it), they change their target by randomly choosing one of the destinations where they think the box exists.

\subsection{\rescue}
At task start, each agent randomly chooses a target location. They will move to the target with probability 1 if there is no additional information. However, if they observe their teammate rescuing people at the place they were targeting, they change the target location by randomly choosing one. If their teammate is at one of the bridges, they will change their destination with 0.1 probability to repair it together.

\subsection{\rescuetwo}
At task start, each agent randomly chooses a target location. They will move to the target with probability 1 if there is no additional information. However, if they observe one of their teammates rescuing people at the place they were targeting, they change the target location by randomly choosing one. If their teammate is at the city hall or mall which requires two agents to rescue people, they will change their destination to there with 0.1 probability even if they were targeting another destination.

\section{Limitations \& Future work}

While we discover numerous factors to consider intervening in a team, our work also leaves several unexplored directions and suggests several areas of future work.
Since we only conducted experiments with simulated teams, the immediate future direction is to explore and model such factors with teams of humans or human-agent. 
Also, we assume the set of mental states is known to derive interpretable and tractable analysis.
It would be worth exploring non-parametric or unsupervised methods to learn mental states and relate their compatibility to team performance.
In addition, the compatibility function (\autoref{eq. cumulative reward}) depends on the cartesian product of state space and mental state space ($U:=S\times X)$. Since $U$ grows exponentially as the number of agents increases, an essential future direction is to examine alternate ways that scale up better to compute this V-value: e.g., via Monte Carlo tree search or state abstraction.
Given that communication is an important part of most teamwork, incorporating communication features into teamwork intervention algorithms is also an important future direction.

% \subsection{Compatible mental models in \movers}
% \label{sec. movers compatible}
% In \movers domain, a member cannot pick up a box alone, two members should think of the same box as their target to complete the task. For each state $s$, we define their mental models are compatible when they think of the same pick-up location as their destination and the box at that location is still there. 

% The acknowledgments are automatically included only in the final and preprint versions of the paper.
\begin{acks}
We thank the anonymous reviewers for their constructive feedback.
This work was supported in part by NSF award $\#2205454$.
Sangwon Seo was partially supported by the Army Research Office through Cooperative Agreement Number W911NF-20-2-0214 and Rice University. 
The views and conclusions contained in this document are those of the authors and should not be interpreted as representing the official policies, either expressed or implied, of the Army Research Office or the U.S. Government.
The U.S. Government is authorized to reproduce and distribute reprints for Government purposes notwithstanding any copyright notation herein.
\end{acks}

% \bibliographystyle{ACM-Reference-Format} 
% \bibliography{main}  % .bib

\end{document}